# Tracing the ongoing emergence of human-like reasoning in Large Language Models

**Paolo Morosi[1*], Nikoleta Pantelidou[1], Fritz Günther[2], Elena Pagliarini[1], Evelina Leivada[1,3]**

1. *Departament de Filologia Catalana, Universitat Autònoma de Barcelona, Barcelona 08193, Spain.*
2. *Institut für Psychologie, Humboldt-Universitat zu Berlin, Berlin 10099, Germany.*
3. *Institució Catalana de Recerca i Estudis Avançats (ICREA), Barcelona 08010, Spain*

*Corresponding author: Paolo Morosi
**Email:** paolo.morosi@uab.cat

## Abstract

Humans effortlessly go beyond literal meanings: “If you mow the lawn, I’ll give you 50$”, is typically understood as implying that the speaker will pay only if the lawn is mowed, whereas “If you’re hungry, there’s pizza in the oven” implies that pizza is available regardless of the hearer’s hunger. Large Language Models (LLMs) show human-like performance on many tasks, yet it remains unclear whether they reason like humans. To address this, we conducted a population-matching experiment assessing how 25 LLMs compute conditional inferences across 4 languages, compared to an equal number of humans per language. We find that humans enrich logical reasoning through pragmatic inferences across languages. Model behavior is more variable. Some LLMs perfectly follow the truth-table of conditionals but they ignore pragmatic inferences, while others deviate from the truth-table, adhering to a single interpretation across the board, thus reflecting accurate rule-based processing but not human-like reasoning. Overall, LLMs are accurate semantic operators, but fail to capture the pragmatic enrichments characteristic of human reasoning. Crucially, LLM accuracy is neither predicted nor boosted by open vs. closed status, training orientation, or architecture type, suggesting that pragmatic reasoning is still an emerging ability in the cognitive toolkit of artificial systems.

## 1. Introduction

Recent advances in Artificial Intelligence (AI), and Large Language Models (LLMs) in particular, have ignited an interdisciplinary debate at the intersection of computer science, cognitive science, and linguistics. The increasingly sophisticated abilities of LLMs have given rise to claims that formal properties of human language have emerged in these systems (1-2). For some scholars, such claims challenge long-standing generative theories of grammar (3-4) and position LLMs as accurate models for studying human learning and processing of language (5). At the same time, it has been argued that LLMs, while undeniably impressive linguistic agents, have cognitive toolkits that remain fundamentally different from those of humans, particularly in their handling of pragmatics and deeper structural generalizations that underpin human linguistic competence (1, 6-9). This debate bears directly on critical questions concerning the nature of human cognition, the limits of AI, and the very definitions of "language" and "understanding" that different disciplines endorse (10-11).

Against this backdrop, a growing body of work in computational linguistics and cognitive science has focused on systematically probing the limits of LLMs' knowledge (e.g., LODNA (12)). This line of research has yielded diverging conclusions. On the one hand, several studies report that LLMs successfully capture important aspects of human language, including morphological generalizations (13-15), polysemy patterns (16), long-distance number agreement (17), and other construction types, such as nominal expressions (18), or target placement in prepositional phrases (19). On the other hand, LLM performance often diverges from human baselines on linguistic phenomena that humans develop early, including core syntactic distinctions (20), explicit reasoning about syntactic structures and rules (21), sensitivity to attested versus unattested word orders (22), and grammaticality judgments across a range of linguistic domains (23-24).

An increasingly growing line of work has argued that, even when LLMs appear to display sophisticated linguistic competence, they nonetheless struggle with meaning-related components of language (8, 25). More specifically, LLMs have been argued to lack language understanding rooted in a perception of real-world reality (26), to exhibit shallow sensitivity to meaning in comprehension tasks (27), and to lack stable conceptual representations and signifiers (28). These limitations entail systematic deviations from human behavior in meaning-sensitive domains, such as passive constructions and negation (20), correlative comparative idiomatic constructions (29) generic statements (30), quantificational expressions (31), and pragmatic implicatures in general (32).

Claiming in broad strokes that LLMs struggle with meaning-related components of language is not sufficient if one cannot indicate the precise source of the challenges. We offer a unifying explanation of these general inconsistencies, arguing that the reasoning abilities of LLMs are affected by what we term a *Decontextualization Bias*: a tendency to rely on formal or literal aspects of linguistic input, without yet fully integrating the contextual cues that guide human processing. One way to frame this distinction is that LLMs have acquired *formal* linguistic competence, but not *functional* linguistic competence (1). This bias gives rise to a systematic preference for literal readings (33), a weakness to interpret implicatures (34) and a tendency to converge on a single interpretation in the face of ambiguous language (35). More generally, given that LLMs do not acquire language grounded in real-world conditions, they may be inherently biased toward literal interpretations at the expense of pragmatically enriched ones. Under this view, although LLMs can potentially capture certain aspects of pragmatic regularities, when a literal interpretation competes with a contextually enriched alternative, models are predicted to diverge from human baselines, as humans exhibit the opposite tendency: they consistently draw pragmatic inferences by adhering to Gricean principles of cooperative communication and by contextualizing incoming stimuli through readily accessible long-term memory resources (36-38).

While limitations in language processing inevitably affect the ability to reason using linguistic input, to determine where exactly the reasoning abilities of LLMs deviate from human baselines, one needs to tap into the complexity of the semantics-pragmatics interface, teasing apart the contribution of the different components behind it. We do that through a novel benchmark that tests how models interpret conditional statements in comparison to humans across 4 languages: Catalan, English, Italian, Spanish. Conditionals constitute a core linguistic construction that integrates semantics, pragmatics, reasoning, and probability assignment (39), making them uniquely suited for disentangling meaning-related components of language in model performance. Specifically, we focus on two types of inferences that humans typically associate with conditionals: *perfected* (40) and *biscuit* (41-42) interpretations. Humans compute both inferences reliably, albeit to different degrees, as discussed below (43-45).

To obtain a broad and representative assessment (46), we evaluated 25 LLMs covering different families and architectures, aggregating model responses at a scale comparable to the human sample, thus facilitating reliable statistical comparison. This broad coverage allows us not only to chart the overall emergence of reasoning in model behavior, but also to investigate which architectural and training choices may boost the development of this property in data-driven systems. Specifically, we focus on 3 largely unexplored main issues.

First, we examine differences between open-source and closed-source models. The distinction is not always straightforward. Open-source systems generally provide greater transparency and accessibility to their training code and data, whereas proprietary models often achieve state-of-the-art performance but restrict access to their architecture and training materials. Importantly, many so-called open-source models provide only open weights, which are downloadable and runnable locally, while their underlying code and training data remain private. Prior work has highlighted that both training data and type of access can influence the capabilities and limitations of LLMs (47). Given that transparency and accessibility increasingly guide both research and deployment decisions, it is important to assess whether model performance systematically varies with open- or closed-sourceness.

Second, we contrast general-purpose generative models with reasoning-oriented models. This comparison is theoretically motivated because it generates two competing predictions. Reasoning-oriented systems, including models optimized for chain-of-thought prompting or tool augmentation (48-49), may be better equipped to integrate key components of reasoning such as logical structure and pragmatic enrichment, benefiting from explicit inference mechanisms. In contrast, general-purpose generative models are trained on large-scale, diverse corpora, which may allow human-like linguistic competence to emerge spontaneously in silico. Comparing models with different training orientations allows us to evaluate whether logical generalizations and pragmatic inferences rely primarily on specialized reasoning mechanisms or can arise naturally in broadly trained, AI generative systems.

Third, we tap into architectural differences by comparing dense models (50-51) with Mixture-of-Experts (MoE) architectures (52-54). This contrast likewise yields competing predictions. MoE architecture substantially increases effective capacity without proportional computational cost and may therefore support richer representations and more flexible inferences. Dense models, by contrast, may promote more stable and globally integrated processing. Comparing these architectures allows us to test whether performance on complex reasoning tasks is primarily shaped by capacity allocation, sparsity, or broader design constraints.

Together, these comparisons enable us to move beyond aggregate accuracy and identify which aspects of model design facilitate — or constrain — the emergence of human-like reasoning in LLMs. We therefore address two main Research Questions (RQs):

RQ1. Do LLMs reason like humans? Specifically, do they exhibit both logical reasoning and pragmatic grounding to the same extent as human speakers when confronted with the same material?

RQ2. Does model performance vary across types of LLMs and languages? Does performance relate to factors such as (i) availability (open- vs. closed-source), (ii) type of training (generative vs. reasoning), and (iii) architectural design (dense vs. MoE)?

## 2. The linguistic properties of conditionals

Conditional statements are chosen to explore the reasoning abilities of LLMs because they are special in combining a semantic component (i.e., the logical meaning of conditionals) with a pragmatic component (i.e., the two inferences). To exemplify, in standard conditionals (SCs) (e.g., "If you mow the lawn, I'll give you 50$"), speakers typically strengthen the sentence to a biconditional, perfected reading (i). In biscuit conditionals (BCs) (e.g., "If you're hungry, there's pizza in the oven"), the consequent is interpreted as holding independently of the antecedent (ii).

(i) If you mow the lawn, I will give you five dollars. (40) *STANDARD*
(ii) If John is hungry, there is pizza in the oven. (apud (41)) *BISCUIT*

In classical propositional logic, however, the SC *if p, then q* is defined as follows:

(iii) $[\![p \rightarrow q]\!]=1$ iff $\neg p \vee q$

On this definition, the conditional is a purely truth-functional operator rather than a causal or temporal one. Therefore, it is true in all cases except when the antecedent *p* is true and the consequent *q* is false. Crucially, when the antecedent is false, the conditional is vacuously true, as the circumstance under which it could be falsified does not arise. Accordingly, in (i), classical semantics allows for the possibility that the hearer receives five dollars even if the lawn is not mowed; for instance, if the hearer tidies their room or cooks dinner.

In natural language interpretation, however, canonical *if*-conditionals typically convey additional information, including causal dependence and relevance between antecedent and consequent, temporal ordering, and speaker expectations (55-56). As a result, in everyday conversation, SCs as (i) are often pragmatically strengthened to a biconditional interpretation (i.e., *q if and only if p*), formally represented in (iv):

(iv) $[\![p \leftrightarrow q]\!] = 1$ iff $([\![p]\!] = [\![q]\!])$

Under this interpretation, the hearer will receive five dollars if and only if they mow the lawn, and not in other circumstances. Importantly, this strengthening does not follow from the truth conditions of the conditional itself but is instead usually taken to be a pragmatic inference (57), commonly modeled as an exhaustivity implicature (58-60). Specifically, exhaustification leads hearers to infer that the stated antecedent exhausts the set of relevant conditions under which the consequent holds, thereby yielding a strengthened, biconditional interpretation.

In contrast, the BC in (ii) differs from SCs (e.g., (i)) in that the consequent *q* is interpreted as holding independently of the antecedent *p*. This interpretation can be schematically characterized as in (v):

(v) $[\![p \rightarrow q]\!] = [\![q]\!]$

Intuitively, in (ii), the existence of pizza in the oven does not depend on John's hunger and holds regardless of whether the antecedent is true or false. Some authors describe this as the "literal meaning" of conditionals (43), insofar as it is compatible with the classical truth table for conditionals. Crucially, however, the biscuit reading does not amount to vacuous truth: the conditional is true not because the antecedent fails, but because antecedent and consequent are understood as causally and epistemically independent.

Under most accounts (59, 61-62), the entailment of the consequent in BCs is instead derived via a pragmatic inference layered on top of standard conditional semantics. Hearers thus infer the epistemic independence of the consequent from the antecedent; namely that neither proposition is informative about the truth of the other relative to a given epistemic state.

The distinct interpretations underlying perfected and biscuit conditionals yield clear empirical predictions about how humans reason regarding the truth of conditional statements across contexts. In the next section, we provide the results of an experiment testing whether the full variation of human interpretive patterns has emerged in LLMs. Our design yields two distinct predictions for each conditional type in the critical conditions, where a pragmatic response is expected. For SCs, participants are predicted to provide a negative response when evaluating the truth of the consequent, consistent with the target pragmatic biconditional inference. In contrast, for BCs, an affirmative response to the same question is predicted, reflecting an inference in which the antecedent and consequent are interpreted as holding independently. In the following section, we provide a detailed description of the experimental design.

## 3. Methods

### 3.1 Experimental design and materials

Building on previous experimental literature on conditionals (43), we designed a truth-value judgment task in 4 languages: Catalan, English, Italian, and Spanish. For each language, we constructed 54 experimental prompts containing either a SC or a BC.

As exemplified in (vi) and (vii), each prompt consisted of (a) a short context specifying the truth of the antecedent *p* and/or the consequent *q*, (b) a brief distractor sentence, and (c) a conditional statement. Participants were then asked to reason about the truth of the consequent *q* on the basis of the conditional uttered in the given context. In addition to the experimental items, we included 18 filler trials (9 true, 9 false) to maintain participant engagement and reduce the transparency of the experimental manipulation.

The 54 experimental items per language included 27 SCs and 27 BCs: 9 critical SCs, 9 true control SCs, 9 false control SCs, 9 critical BCs, 9 true control BCs, 9 false control BCs. These items were distributed across three lists following a Latin-square design, such that each participant saw each item in only one condition. Each list therefore contained 27 trials: 9 SCs, 9 BCs, and 9 fillers (5 true and 4 false). SC and BC items appeared in three truth-table configurations:

A. **True control conditions** (3 items per conditional type): the antecedent and consequent were both true in the context ($p \wedge q$), and the consequent was therefore expected to be judged true.
B. **False control conditions** (3 items per conditional type): the antecedent was true but the consequent was false ($p \wedge \neg q$), and the consequent was therefore expected to be judged false.
C. **Critical conditions** (3 items per conditional type): the antecedent was false ($\neg p \wedge q?$) In these cases, the expected judgment diverged by conditional type: the consequent was expected to be judged false for SCs (reflecting a perfected, biconditional reading) and true for BCs (reflecting a biscuit reading and the inference that *p* and *q* are independent). In other words, for SCs the target reading is the one that tightly links the truth value of the consequent to that of the antecedent, whereas in BCs, the two are expected to be interpreted as independent.

Examples of critical SC and BC trials are given in (vi) and (vii), respectively.

(vi) Here's the context: The lawn has not been mowed. Also, the flowers still need watering. Mary had said: "If Paul mows the lawn, he will get 50€". Based on what Mary says, is it true that Paul got 50€?

(vii) Here's the context: John is not hungry. There is pasta in the fridge. Mary says: "If John is hungry, there is pizza in the oven". Based on what Mary says, is it true that there is pizza in the oven?

The target "no" response in (vi) indicates a perfected biconditional interpretation, according to which Paul would receive €50 if and only if he mows the lawn. In contrast, the target "yes" response to the BC in (vii) indicates that there is pizza in the oven regardless of John's hunger, suggesting that the antecedent and consequent are independent. The full set of prompts is publicly available at https://osf.io/xuqbw/overview.

### 3.2 Human Participants

We collected truth-value judgments from a total of 364 participants (F=162, non-binary=4), recruited via the crowdsourcing platform Prolific. All participants were self-identified native speakers of the target language (Catalan, English, Italian, or Spanish).

Participants were excluded if they met at least one of the following criteria: (i) self-reported speech- or language-related disorders (e.g., dyslexia), (ii) prior or ongoing treatment with a speech/language pathologist, (iii) self-identification as neurodivergent (e.g., autism), or (iv) failure to correctly respond to at least 6 out of the 9 filler items. After exclusions, the final sample comprised 313 participants (F=136, mean age: 38;15). See Table 1 for the sample size and demographic information of participants.

| | N | Age (Mean) | Female | Male | Non-binary |
|---|---|---|---|---|---|
| Catalan | 81 | 33;64 | 40 | 40 | 1 |
| English | 78 | 45;83 | 46 | 32 | 0 |
| Italian | 76 | 35;25 | 28 | 48 | 0 |
| Spanish | 78 | 37;86 | 22 | 55 | 1 |
| **Total** | **313** | **38;15** | **136** | **175** | **2** |

**Table 1.** Sample size and demographic information of the participants included in the analysis.

For each language, participants were evenly assigned to one of three lists following the Latin-square design, resulting in at least 25 participants per list.

Trials were pseudo-randomized to prevent consecutive presentation of items of the same conditional type or condition. After providing informed consent and demographic and sociolinguistic information, participants were instructed that they would read short texts and answer a question on each trial, and that they would not be able to return to previous trials, once they advanced.

On each trial, participants saw the full prompt on the screen together with a text box in which they typed their response and a "Next" button to proceed. The prompt remained visible while participants typed their answer. Once the response was submitted, the trial could not be re-accessed. Responses were manually mapped to trinary truth judgments (true/false/undetermined) and subsequently scored for accuracy as 1 (correct) or 0 (incorrect) according to the target response.

The experiment was implemented using the PCIbex Farm platform (63). The median completion time was 19.23 minutes. Anonymized participant data is available at the OSF repository. The study was conducted in accordance with the Declaration of Helsinki and approved by the Ethics Committee for Research (*Comité d'Ètica en la Recerca*, CERec) of the Autonomous University of Barcelona (application CEEAH 7150).

### 3.3 LLMs

We evaluated 25 LLMs encompassing both commercial and open-source systems, dense and MoE architectures, and generative as well as reasoning models, including representatives from the Claude, GPT, Gemini, Llama, Mistral, Falcon, and Qwen families. A complete list of tested models, along with their interface, open-source status, architecture, and model type, is provided in Table 2.

| LLM | Interface | Availability | Architecture | Type |
|---|---|---|---|---|
| Claude Haiku 4.5 (65) | Open Router | Closed | Dense | Generative (Hybrid) |
| Claude Opus 4 (65) | Open Router | Closed | Dense | Reasoning |
| Claude Sonnet 4.5 (66) | Open Router | Closed | Dense | Generative (Hybrid) |
| DeepSeek-v3.1 (67) | Ollama | Open weight | MoE | Generative (Hybrid) |
| Distill-bert (Distill-distilbert-base-cased-distilled-squad) (68) | Hugging Face | Open source | Dense | Generative |
| Falcon 7-B (69) | Hugging Face | Open source | Dense | Generative |
| Gemini 2.5 Pro (70) | Open Router | Closed | MoE | Reasoning |
| Gemini 2.5 Flash (70) | Open Router | Closed | MoE | Reasoning |
| Gemma-3-4b-it (71) | Open Router | Open weight | Dense | Generative |
| GLM-4.6 (72) | Ollama | Open source | MoE | Reasoning (Hybrid) |
| GPT-4o (73) | Open Router | Closed | Dense | Generative |
| GPT-4o-mini (73) | Open Router | Closed | Dense | Reasoning |
| GPT-5 (74) | Open Router Free | Closed | Dense | Reasoning |
| GPT-oss 120b (74) | Ollama | Open weight | MoE | Reasoning |

| Grok 4 Fast (75) | Open Router | Closed | Dense | Generative (Hybrid) |
|---|---|---|---|---|
| Kimi K2-Instruct-0905 (76) | Ollama | Open source | MoE | Generative |
| Llama3.3 70 b Instruct (77) | Open Router Free | Open source | Dense | Generative |
| Llama4 Scout (78) | Open Router Free | Open weight | MoE | Generative |
| Mistral-7B Instruct v0.2 (79) | Hugging Face | Open source | Dense | Generative |
| Nvidia-nemotron-nano-9b-v2 (80) | Open Router | Open source | Dense | Reasoning (Hybrid) |
| OLMo 2-0325-32b-instruct (81) | Open Router | Open source | Dense | Generative |
| Perplexity sonar-reasoning-pro (82) | Open Router | Closed | Dense | Reasoning |
| Phi-4 reasoning plus (83) | Open Router | Open weight | Dense | Reasoning |
| Qwen3-vl-235b (84) | Ollama | Open weight | MoE | Reasoning |
| Virtuoso Large (85) | Open Router | Closed | Dense | Generative |

**Table 2.** Characteristics of the 25 tested models.

All models were tested on the same stimuli used with human participants. For each language, models were presented with the complete set of 54 experimental items and 18 fillers (i.e., all three Latin-square lists), ensuring identical coverage of conditions across systems. Using the full dataset per language allowed us to obtain 72 judgments per model (54 experimental + 18 fillers) and to aggregate responses across models to match the scale of the human sample for statistical comparison. The model responses were collected in two sessions over the course of various days.

Prompts were identical to those shown to humans (e.g., (vi)-(vii)) and were submitted verbatim without additional instructions beyond the task question. Models generated free-text responses, which were manually mapped to binary truth judgments (true/false) and scored for accuracy as 1 (correct) or 0 (incorrect) based on the target response, consistent with the scoring procedure used for human participants. All the tested LLMs were queried programmatically using Python 3.12 (86). As shown in Table 2, access was handled through 3 AI model runners: Hugging Face (87), Ollama (88), and OpenRouter (89). Testing was conducted between October and November 2025.

All the prompts, the anonymized results, and the code for the statistical analyses are available at https://osf.io/xuqbw/overview.

## 4. Results and analysis

We compared human participants and LLMs on a truth-value judgment task designed to dissociate logical from pragmatic interpretations of conditionals, focusing on the distinction between perfected and biscuit readings. The primary dependent measure was binary accuracy (i.e. correct vs. incorrect interpretations relative to the target answer; see Methods for further details).

We analyzed the data with logistic regression analyses, using generalized linear-mixed effects models in R to account for the nested data structure (90-91). Our baseline model contained only an intercept, plus random intercepts for participants and items. In this analysis, each LLM is treated equivalent to an individual human participant. Removing any of these random intercepts resulted in a significant decrease of model fit in a likelihood-ratio test ($X^2(1) = 1264$, $p < .001$ and $X^2(1) = 2143.8$, $p < .001$, respectively). We then identified our statistical model via forward model selection with likelihood-ratio tests, successively entering additional predictors in a theoretically motivated order. The results of this model selection process are displayed in Table 3.

| Step | Additional predictors | $X^2$ | df | *p* |
|---|---|---|---|---|
| 0 | 1 + (1\|participant) + (1\|item) | | | |
| 1 | + Conditional [Standard vs. Biscuit] | 0.03 | 1 | .856 |
| 2 | + Prompt [Experimental vs. Control] | 38.74 | 1 | < .001 |
| 3 | + Prompt*Conditional | 33.70 | 1 | < .001 |
| 4 | + Agent [human vs. LLM] | 25.73 | 1 | < .001 |
| 5 | + Prompt*Agent | 762.85 | 1 | < .001 |
| **Model parameters of the resulting model (Step 5)** | | | | |
| | Fixed effect | *b* | *z* | *p* |
| | Intercept [Control, Biscuit, human] | 2.24 | 12.56 | < .001 |
| | Prompt [Experimental] | -1.88 | -6.60 | < .001 |
| | Conditional [Standard] | -0.77 | -3.33 | < .001 |
| | Agent [LLM] | 0.13 | 0.61 | .540 |
| | Prompt [Experimental] * Condition [Standard] | 3.07 | 7.67 | < .001 |
| | Prompt [Experimental] * Agent [LLM] | -3.19 | -25.75 | < .001 |

**Table 3.** Upper part. Results of the likelihood ratio tests for the logistic mixed-effects models with the binary variable accuracy as the dependent variable (1: correct, 0: incorrect). Predictors that did not significantly improve model fit (marked in grey) were not included in the model for the subsequent steps.

The resulting model in Table 3 revealed significant effects of prompt type and its interactions with both conditional type and agent. In particular, the Prompt*Conditional interaction indicates that the differences between Standard and Biscuit conditionals is substantially stronger in the experimental trials, whereas the Prompt*Agent interaction shows that the experimental manipulation differentially affected human participants and LLMs (see also Figure 1).

## 4.1 Agent comparison

Across all languages, humans performed near ceiling in the true control conditions for both conditional types but showed lower accuracy in false control conditions. In the experimental condition, human accuracy is higher for SCs than for BCs. Comparing humans and LLMs, we observe no differences in the control conditions. In the experimental conditions, however, LLMs perform significantly worse than human participants. We therefore conducted a focused analysis on experimental trials only.

In this subset of the data, including an interaction between Agent and Conditional —in addition to their respective main effects and the intercepts for participants and items [Accuracy ~ Agent + Conditional + (1 | participant) + (1 | item)]— does not significantly improve model fit ($X^2(1) = 0.116$, p = .733). As can be seen in Figure 1, in the experimental conditions we observe the same difference in accuracy between BCs and SCs (higher for Standard; $b = 2.68$, $z = 7.68$, p < .001) in humans and LLMs, but at an overall lower level for LLMs ($b = -3.50$, $z = -10.83$, p < .001). Overall, human accuracy remained relatively stable across control and experimental trials, whereas LLM accuracy dropped substantially in experimental contexts requiring pragmatic reasoning (Figure 1). Thus, although LLMs showed some sensitivity to the different conditions, behaving as reliable logical agents, their responses were markedly less consistent in critical conditions, which required a pragmatic interpretation.

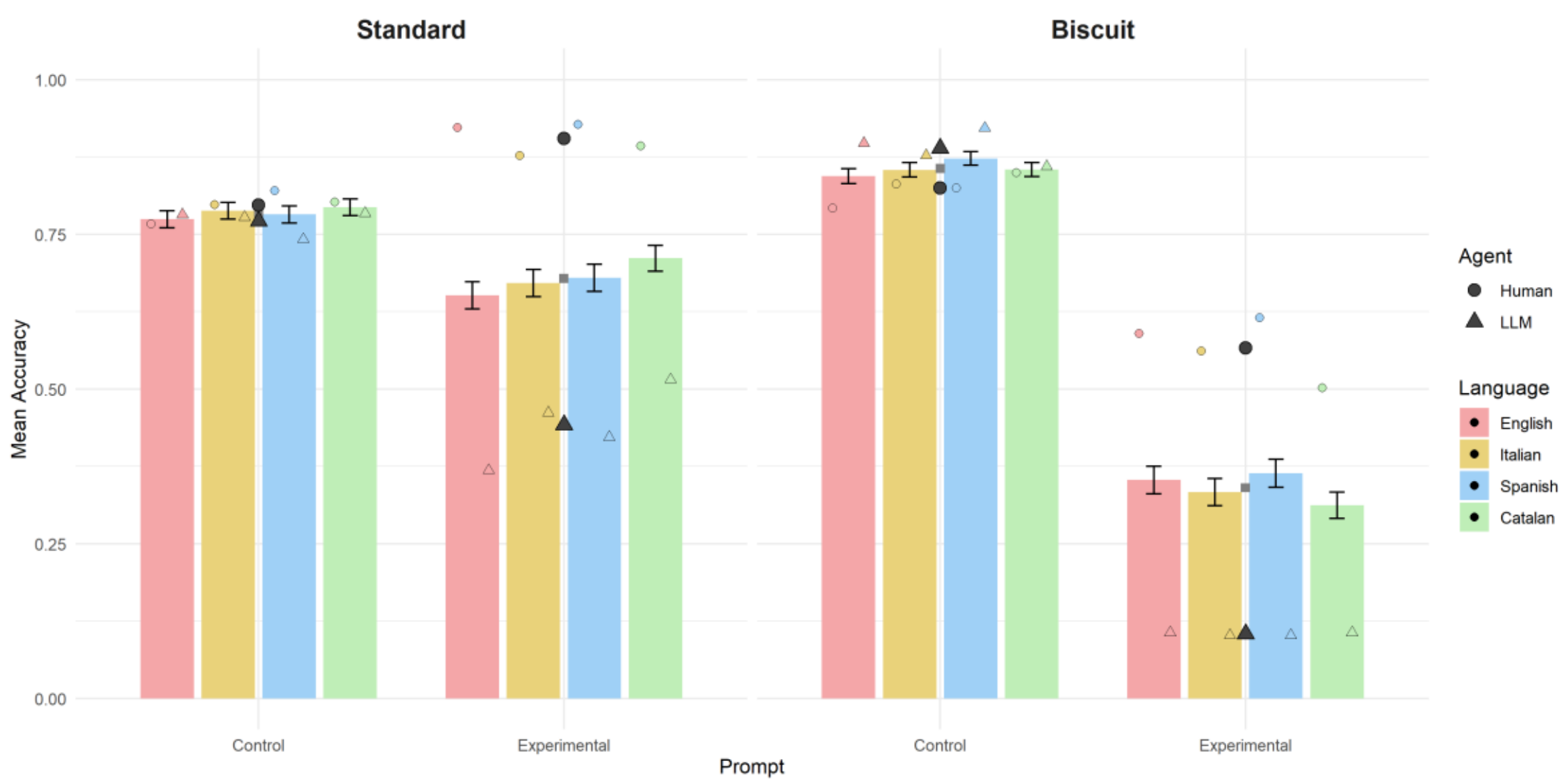

**Figure 1.** Accuracy across conditions, agents, and languages. Colored triangles indicate mean LLM accuracy per language, while colored circles indicate mean human accuracy per language. Black triangles and circles indicate overall accuracy for models and humans respectively. Vertical bars represent SE.

## 4.2 Models and languages comparison

Next, we turned to differences between individual LLM architectures. Within this subset of the data that only contains LLM responses, removing the random intercept for "participants" from the described model leads to a significant decrease in model fit ($X^2(1)$

= 915.86, p < .001), indicating high variation between different LLM architectures in terms of their accuracy in the task at hand. As Figure 2 shows, mean accuracies range from .000 (Distill-Bert) to .833 (Llama3.3). Overall, substantial heterogeneity emerged across individual models. Beyond general accuracy, models differed in their sensitivity to conditional type. Allowing the Standard–Biscuit contrast to vary by model significantly improved fit ($\chi^2$ = 106.04, p < .001), demonstrating that some models reliably distinguished the two conditional types whereas others did not. For instance, GPT-4o-mini and Grok 4 Fast exhibited strong differentiation, with differences exceeding 0.6 between Standard and Biscuit in experimental trials, whereas models such as Falcon 7-B or GPT5 showed weak or even reversed contrasts.

We then tested whether LLM performance differs across different languages. Starting from a model containing as fixed effect the interaction between Prompt and Conditional, plus random intercepts for participants and items [Accuracy ~ Prompt*Conditional + (1 | participant) + (1 | item)], we tested whether additionally including a random intercept for languages improved model fit. This was not the case ($X^2(1) = 0.000$, p > .999), with very similar mean accuracy being observed in all 4 tested languages (ranging from .639 in English to .652 in Catalan). Conditional interpretation therefore appears largely language-independent within the tested set of languages, suggesting that pragmatic limitations arise primarily from model-internal factors rather than language-specific processing constraints. We also observed modest model × language interactions ($\chi^2$ = 12.76, p < .001), indicating that some models' ability to generalize pragmatic distinctions could be influenced by language, although these effects were far smaller than model-specific baseline differences (Figure 3).

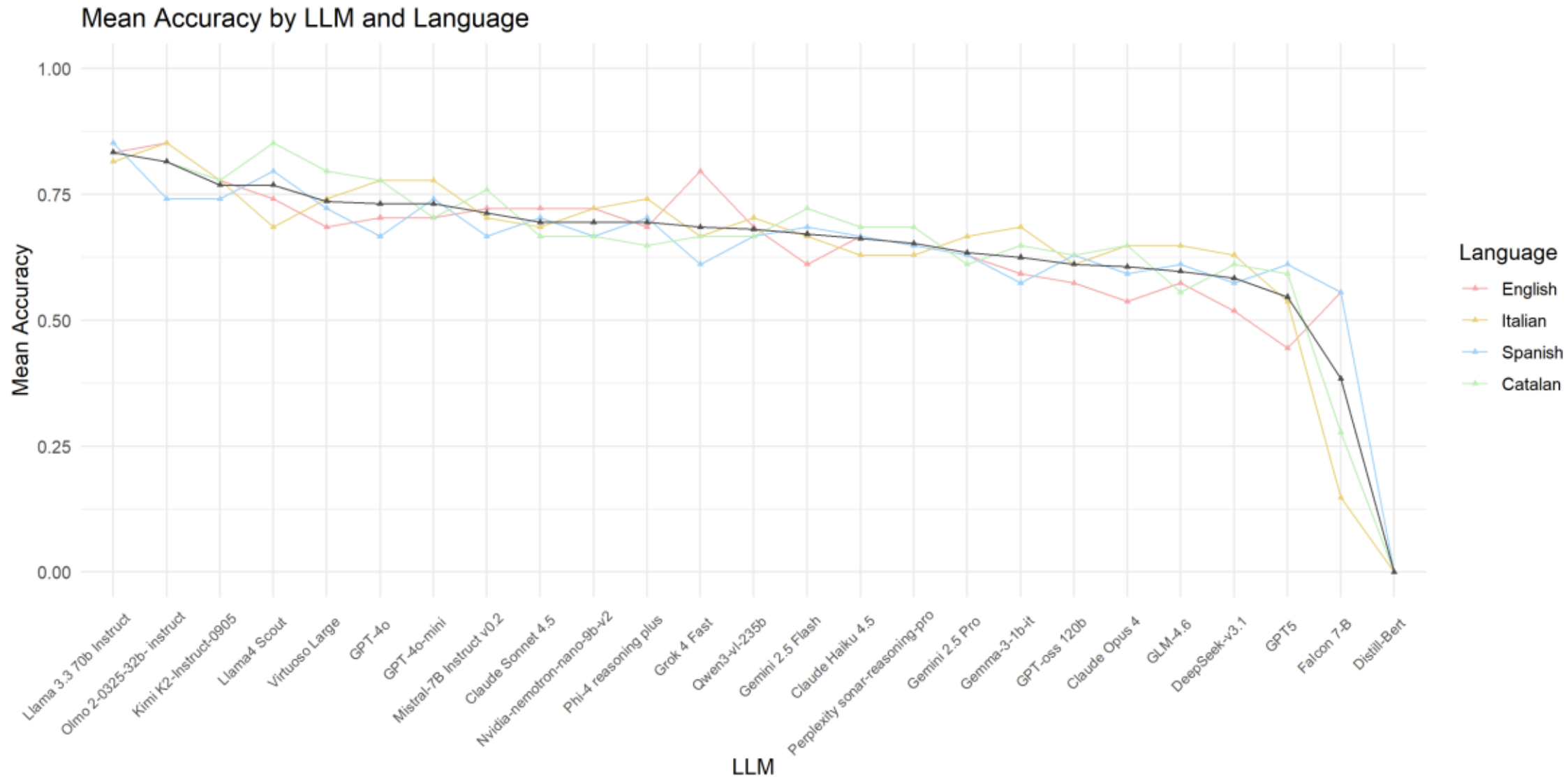

**Figure 2.** Mean accuracy in experimental trials by model and language. The black line represents model mean accuracy across languages.

We further tested whether LLMs differ in their sensitivity to the SCs and BCs contrast. Including a by-participant random slope of Conditionals [Accuracy ~ Prompt*Conditional + (Conditional | participant) + (1 | item) + (1|language) +

(1|participant: language)] significantly increased model fit ($X^2(2) = 106.04$, $p < .001$), confirming that models vary widely in how strongly they differentiate between conditional types. Some models had a much better performance for SCs (e.g., Llama3.3: accuracy .972 for SCs, and .167 for BCs), some models —all with very low overall performance— showed only a small difference (e.g., Gemini 2.5 Pro: accuracy .111 for SCs and .028 for BCs), and one model even showed the reverse pattern (Falcon 7-B: accuracy .056 for SCs and .472 for BCs). Overall, we observe great heterogeneity in model performance. Some high-performing models, such as Llama3.3 70b Instruct and Kimi K2-Instruct-0905, maintained accuracies above 0.75 across languages, while other models, such as DeepSeek-v3.1 or GPT5, remained near chance regardless of language (Figure 3).

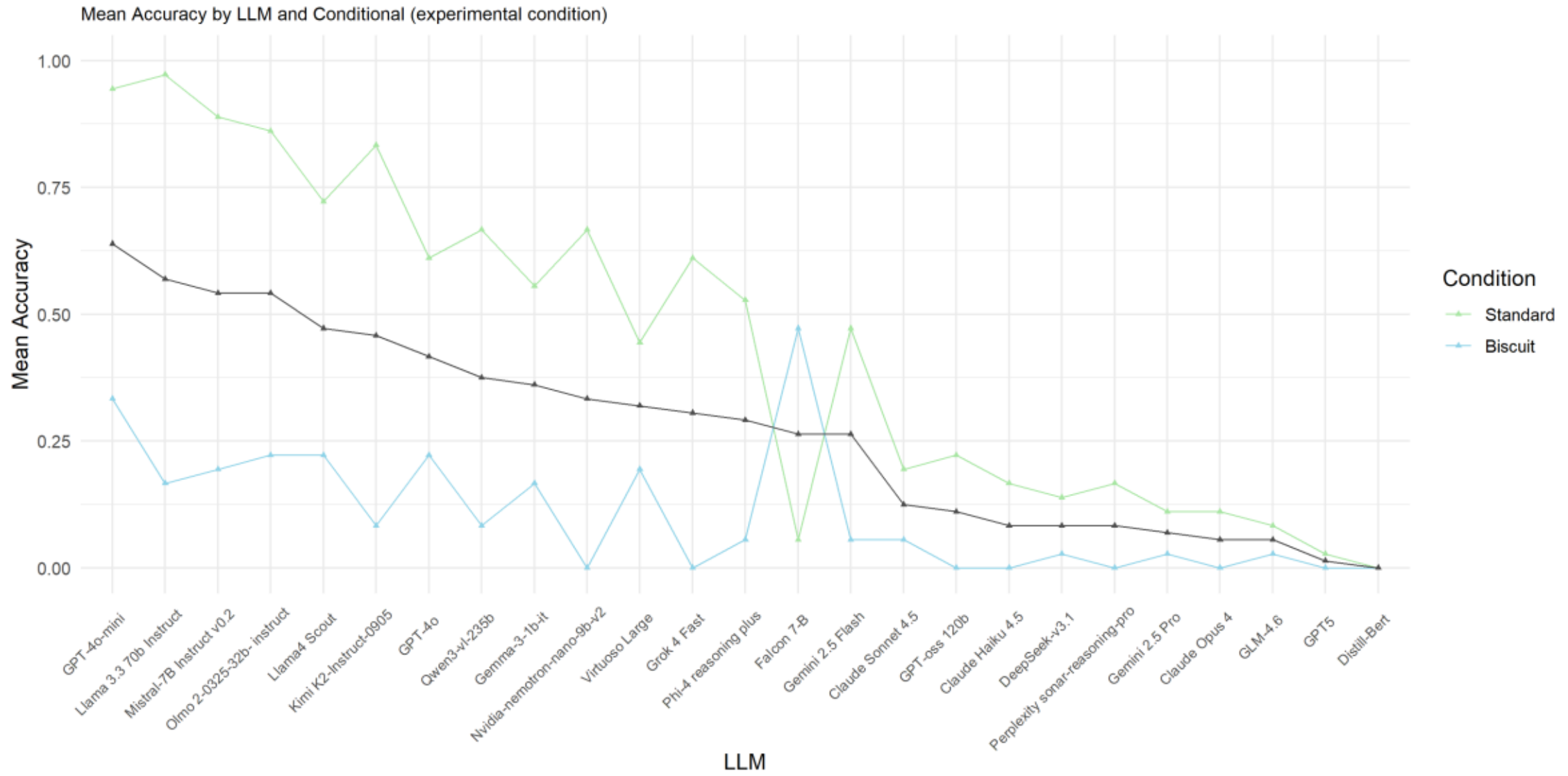

**Figure 3.** Mean accuracy across languages for all models in critical Standard and critical Biscuit conditions. The black line represents mean accuracy across the two conditions.

Finally, we examined whether broad design features could account for the observed variation across models. Specifically, we tested 3 factors: openness (open vs. closed models), type of architecture (dense vs. mixture-of-experts), and training orientation (generative vs. reasoning-oriented). Crucially, none of these factors reliably predicted better model performance. Including openness as a fixed effect did not lead to a significant increase in model fit ($X^2(1) = 0.307$, $p < .580$), presenting no evidence for general differences in accuracy between open-source and proprietary models (Table 4). Similarly, adding architectural type ($\chi^2 = 0.15$, $p = .70$) or model type (generative vs. reasoning-oriented; $\chi^2 = 0.01$, $p = .996$) did not significantly improve fit. Mean differences across these categories were small and statistically indistinguishable, suggesting that high-level design choices do not systematically accuracy in our reasoning task. Instead, performance differences appear to be largely model-specific, reflecting idiosyncratic and largely opaque combinations of training data, internal representations, and fine-tuning strategies. For example, while two dense models might differ dramatically in overall accuracy and pragmatic differentiation, a mixture-of-experts (MoE) system could outperform them in some languages or conditions but not others. These findings underscore that the computation of pragmatic inferences in LLMs is highly contingent on the internal properties of individual models, rather than being determined by broad architectural or training categories (Table 4).

| **Feature** | **Category** | **Mean Accuracy** |
|---|---|---|
| Open status | Open | 0.626 |
| | Closed | 0.668 |
| Architecture | Dense | 0.636 |
| | MoE | 0.664 |
| Training type | Generative | 0.638 |
| | Hybrid | 0.653 |
| | Reasoning | 0.648 |

**Table 4.** Mean accuracy by design characteristics. No category demonstrated a statistically significant effect on overall accuracy (all $\chi^2 < 1$, $p > 0.5$).

## 5. Discussion

This experiment investigates whether LLMs reason in ways comparable to humans. Across 4 languages, 313 humans, and 25 state-of-the-art LLMs, we observe a consistent pattern: humans exhibit robust, context-sensitive pragmatic enrichment, whereas models are reliable logical operators but show only partial and highly variable sensitivity to these inferences. Overall, our findings suggest that the tested LLMs successfully capture the truth-conditional semantics of conditionals but struggle to compute the pragmatic reasoning that characterizes human interpretation. Below, we discuss these findings in relation to our two main Research Questions.

With respect to our primary Research Question —whether LLMs reason like humans, interpreting conditionals in a human-like manner— the human data replicated and extended earlier findings from English-speaking populations (43), using a different methodology and testing across different languages. Evcen & Barner report that speakers derive perfected interpretations of SCs in approximately 82% of cases, whereas BCs receive biscuit interpretations in approximately 58% of cases (43). In our study, humans perfected SCs at a stable rate of more than 88% on average across languages, while biscuit readings for BCs were observed at 57% on average. Thus, human participants reliably distinguished SCs from BCs: perfected interpretations were endorsed consistently, while biscuit interpretations were derived only about half of the times.

LLMs exhibited some sensitivity to conditional structure, but their performance sharply deviated from human baselines in pragmatically demanding contexts (Figure 1): in critical trials, models rarely produced biscuit interpretations and only inconsistently produced perfected readings. Thus, even when models reasoned in a way that captured the human distinction in inferences, distinguishing control from experimental conditions and SCs from BCs, their responses did not fully match human baselines in terms of accuracy.

Inspection of individual model performance revealed two broad behavioral profiles. A first group of models (i.e., Perplexity sonar-reasoning-pro, GPT-4o, GPT-5, GPT-oss 120b, DeepSeek-v3.1, Claude Sonnet 4.5, Claude Haiku 4.5, Claude Opus 4, Qwen3-vl-235b, Gemini 2.5 Pro, Virtuoso Large, GLM-4.6) adhered closely to the truth-table semantics of conditionals, treating sentences with false antecedents as vacuously true and consequently failing whenever contextual reasoning about the consequent was required (i.e., in critical conditions). A second group (i.e., OLMo 2-0325-32b-instruct, Nvidia-nemotron-nano-9b-v2, Kimi K2-Instruct-0905, GPT-4o-mini, Llama3.3 70 b Instruct, Grok 4 Fast, Mistral-7B Instruct v0.2, Llama4 Scout) instead adopted a biconditional interpretation, strengthening *if p, q* to *q if and only if p* across both conditional types. This behavior implies that these models do not engage in genuine pragmatic reasoning, but rather rely on a fixed, rule-based strategy that treats conditionals as biconditionals across the board. Importantly, although this strategy yields superficially good performance on perfected standard conditionals, it systematically precludes a biscuit interpretation. Some models deviated from these two dominant patterns. Gemini 2.5 Flash, Gemma-3-4b-it, Phi-4 reasoning plus, for instance, showed a more variable behavior, generally avoiding biscuit readings in critical BCs, but occasionally producing biconditional interpretations for SCs. Falcon 7-B tended to provide positive responses across conditions and conditional types, suggesting a positive response bias.

Nonetheless, the two general response profiles displayed by LLMs explain the overall distribution of results: biscuit interpretations were nearly absent, whereas accuracy on SCs was comparatively higher. Taken together, these findings suggest that the tested LLMs behave more like competent semantic agents rather than pragmatic reasoners. While models often capture the literal, truth-conditional structure of conditionals, they struggle to integrate contextual cues and speaker intentions in the flexible, inference-driven manner characteristic of human interpretation. This pattern is consistent with what we term a *Decontextualization Bias*: a tendency to privilege surface form and literal meaning over contextually enriched interpretations, which stems from absence of grounding language in real-world communicative settings. Under this view, models may acquire certain pragmatic regularities, but when literal and enriched interpretations *compete*, they resort to the former, diverging systematically from human behavior. This aligns with the observation that LLMs appear to have developed *formal* linguistic competence without fully attaining *functional* competence, defined as the ability to use language in contextually and world-grounded ways (1).

Turning to the interpretive differences between perfected and biscuit readings, human behavior clearly supports this distinction: the two constructions diverged in contexts where a pragmatic inference was required, consistent with accounts in which strengthened readings arise through implicature calculation (58-60), whereas biscuit interpretations depend on reasoning about epistemic independence between antecedent and consequent (59, 61-62, 92).

Crucially, LLM behavior mirrors this distinction only weakly. Although some models exhibited a reduced SC vs. BC contrast, the effect was attenuated and highly inconsistent across models. Many models produced either literal, truth-conditional interpretations or uniformly strengthened biconditionals, thereby collapsing the pragmatic distinction between SCs and BCs observed in human judgments. Specifically, rather than flexibly computing different inferences depending on context, models often applied a single interpretive strategy across constructions, as independently observed in previous work (33, 35). These results reinforce the idea that pragmatic enrichment is not a uniform or automatic consequence of large-scale language modeling. Instead, it appears to require mechanisms for context integration and relevance reasoning that are only partially, if at all, captured by current architectures. Overall, we find that reasoning can be conceived as a multi-step process that integrates various components (e.g., temporal sequencing, causal linking, logical processing, pragmatic enrichment, world-based inference), and at present, the outermost layers of this multi-step process, that are responsible for rewriting the interpretability of the linguistic message based on pragmatic, world-grounded inferences, are still under development in LLMs.

Our second Research Question examined whether differences in LLM performance could be predicted by language or broad design choices. Language appeared to play little role: both humans and models showed highly similar patterns across Catalan, English, Italian, and Spanish. This suggests that conditional reasoning in this task is largely language-independent and that cross-linguistic transfer is not a primary bottleneck for current systems.

Nonetheless, closer inspection of model responses revealed a high degree of cross-linguistic "leakage". Specifically, some models (e.g., GPT-4o, Grok 4 Fast, Mistral-7B Instruct v0.2, Falcon 7-B, Nvidia-nemotron-nano-9b-v2) produced outputs in English — and occasionally in Spanish and other languages — when queried in languages distinct from English. This effect was particularly pronounced for Catalan, the language spoken by the smaller speaker community from the ones we tested, where responses were often produced in a linguistically similar language such as Spanish, or in English. This may suggest that, for these models, English remains the dominant and primary operational language (93-96). Therefore, although model performance was relatively uniform in the present task, a deeper examination of model performance through further larger-scale testing may reveal important cross-linguistic disparities.

In contrast, substantial variability emerged across individual models. Baseline accuracy and sensitivity to conditional type differed widely: some systems approximated human accuracy (e.g., Kimi K2-Instruct-0905, Llama 3.3 70B Instruct, Llama 4 Scout, OLMo 2-0325-32B-Instruct), whereas others performed near or below chance (e.g., Falcon 7B, GLM-4.6, GPT-5), with one model performing at floor levels (Figures 2 and 3). While as an overall class, LLMs do not align with human responses when it comes to reasoning, we find that a few models successfully approximate human baselines. At the same time, this high performance still deviates from human behavior in two ways. First, unlike humans, LLM performance was largely rigid *across conditions*, reflecting rule-based interpretation (i.e., either all conditionals are biconditionals, or all conditionals follow the truth table), and not pragmatically enriched reasoning. Second, LLM performance was highly heterogenous *across models*. Crucially, this heterogeneity was not explained by broad design characteristics, and such that we cannot claim that the best-performing LLMs reason in a human-like way as a direct consequence of specific aspects of their status, training, or architecture. Accuracy did not systematically track openness (open vs. closed source), architectural sparsity (dense vs. mixture-of-experts), or training orientation (generative vs. reasoning-focused). Both high- and low-performing systems were found within each category. In other words, coarse-grained taxonomic labels failed to predict model accuracy in the given task. This finding has practical and theoretical implications. From an engineering perspective, it suggests that high-level design labels are poor predictors of reasoning competence, when this requires pragmatic enrichment. From a cognitive perspective, it indicates that human-like interpretation is not an automatic byproduct of scale or architectural sophistication, but rather an emergent, model-specific property that depends on details not captured by coarse taxonomies at present.

Several limitations to this study should be noted. For instance, although we tested 4 typologically related languages and observed highly consistent patterns across them, our sample remains limited; broader cross-linguistic coverage, including non-Indo-European and lower-resource languages, would be necessary to fully assess the generalization of these findings. Moreover, while we compared models spanning diverse families and design choices, the distribution of intrinsic characteristics (e.g., open vs. closed source, dense vs. MoE, generative vs. reasoning-oriented training) was not perfectly balanced, limiting strong causal inferences about how specific architectural or training factors may genuinely shape pragmatic competence. Last, our evaluation relied on controlled, sentence-level judgments that isolate semantic and pragmatic reasoning but may under- or overestimate performance in richer, discourse-level tasks where additional contextual cues are available.

To conclude, this study examined how LLMs interpret conditional statements in comparison to human baselines. Overall, our results reveal a systematic gap between semantic and pragmatic competence in current models. While LLMs frequently reproduce the formal logic of conditionals, they fail to consistently perform the contextual reasoning that humans deploy naturally. This pattern aligns with the idea that LLMs exhibit a Decontextualization Bias, that favors surface form and statistical regularities over deeper integration of discourse context and communicative intents. Importantly, these findings do not imply that pragmatic reasoning is inherently unattainable for LLMs. Rather, they suggest that such abilities do not emerge automatically from scale alone. Achieving human-like reasoning may require additional mechanisms, including richer grounding in the world, interactive learning environments, or explicit modeling of speaker goals and intentions.

**Ethical statement**

The study was conducted in accordance with the Declaration of Helsinki and approved by the Ethics Committee for Research (*Comité d'Ètica en la Recerca*, CERec) of the Autonomous University of Barcelona (application CEEAH 7150). All human participants gave informed consent to participate in the study.

**Funding statement**
EL acknowledges funding from the Spanish Ministry of Science & Innovation under the research project CNS2023-144415. EL and FG acknowledge joint funding from the Spanish Ministry of Science & Innovation and the German Research Foundation under the research project PID2024-157681NI-I00. F.G. acknowledges funding from the German Research Foundation (Deutsche Forschungsgemeinschaft) under the Emmy-Noether grant "What's in a name?" (project No. 459717703). EP acknowledges the financial support of Grant PID2022-138413NB-I00, funded by MCIN/AEI/10.13039/501100011033 and by the European Union and of Grant RYC2021-033969-I (Ramón y Cajal), funded by MCIN/AEI/10.13039/501100011033 and by European Union Next-GenerationEU/PRTR.

**Data accessibility**
All materials associated with this study are publicly available on the OSF repository at https://osf.io/xuqbw/overview. This repository includes the full set of experimental prompts, anonymized dataset, and the code used for statistical analyses, as well as relevant documentation to reproduce the results.

**Competing interests**
The authors declare no competing interests.

**Authors' contributions**
PM, EP, and EL designed research; PM and NP performed research; NP and FG analyzed data; PM wrote the paper; all the other authors edited, revised, and approved the paper.